\documentclass[letterpaper]{article}
\usepackage{aaai}
\usepackage{times}
\usepackage{helvet}
\usepackage{courier}
\usepackage{amsmath}
\usepackage{xcolor}
\usepackage{multirow}

\usepackage{graphicx}
\usepackage{url}
\usepackage{natbib}  
\usepackage{caption} 
\usepackage{algorithm}
\usepackage{algorithmic}
\usepackage{bibentry}
\usepackage{newfloat}
\usepackage{listings}

\lstdefinestyle{llmprompt}{%
basicstyle=\ttfamily\normalsize,
columns=fullflexible,
backgroundcolor=\color{gray!5},
frame=none,
framerule=0pt,
framesep=0pt,
breaklines=true,
keywordstyle=\bfseries,
keywordstyle=[1]\color{black!70!black},     
keywordstyle=[2]\color{black!50!black},    
keywordstyle=[3]\color{black!60!black},   
keywordstyle=[4]\color{black!70!black},   
morekeywords=[1]{system},
morekeywords=[2]{user},
morekeywords=[3]{assistant},
morekeywords=[4]{tool},
numbers=left,
numberstyle=\tiny\color{gray!70},
numbersep=5pt,
stepnumber=1,
firstnumber=1,
numberblanklines=true
}

\begin{document}
%
\title{Boosting Instruction Following at Scale}
\author{Ben Elder, Evelyn Duesterwald, Vinod Muthusamy\\
IBM T.J. Watson Research\\
Yorktown Heights, NY\\
}

\maketitle
\begin{abstract}
\begin{quote}
A typical approach developers follow to influence
an LLM’s behavior in an application is through careful manipulation of the
prompt, such as by adding or modifying instructions. However, merely adding more instructions provides little assurance that they will actually be followed.  We introduce Instruction Boosting as a post-generation method to increase the reliability of LLM prompt instructions.  
We show that Instruction Boosting improves the instruction following rate by up to 7 points for two instructions and up to 4 points for ten instructions. To demonstrate these results we introduce \textsc{ScaledIF},  a benchmark with a scaled instruction volume of up to ten instructions per data sample. We also present an analysis of the commonly observed trend that performance degrades as more instructions are added. We show that an important factor contributing to this trend is the degree of tension and conflict that arises as the number of instructions is increased. We contribute a quantitative conflict scoring tool that explains the observed performance trends and provides feedback to developers on the impact that additional prompt instructions have on a model's performance. 
\end{quote}
\end{abstract}

\section{Introduction}

Large Language Models (LLMs) have become foundational components in the development of agentic applications. However, for developers, these powerful models often behave like black boxes, making it difficult to precisely control their output. A typical method for influencing an LLM's behavior is through careful manipulation of the prompt, such as by adding or modifying instructions. For instance, when testing reveals an undesirable model outcome, a developer might add a corrective instruction to the prompt in an effort to prevent that behavior from recurring. 

This reliance on prompt-based instructions, however, presents two fundamental problems. First, there is little guarantee that a newly added instruction in the prompt will actually be followed by the LLM. Second, the progressive addition of instructions can inadvertently introduce tension or even direct contradictions with pre-existing instructions, making it more difficult to satisfy all instructions simultaneously. These issues combined can lead to an often observed phenomenon, where the instruction following rate (IF rate) degrades as the number of instructions increases~\citep{jiang2024followbench}.

\begin{figure}[t]
    \centering
    \includegraphics[width=1.0\linewidth]{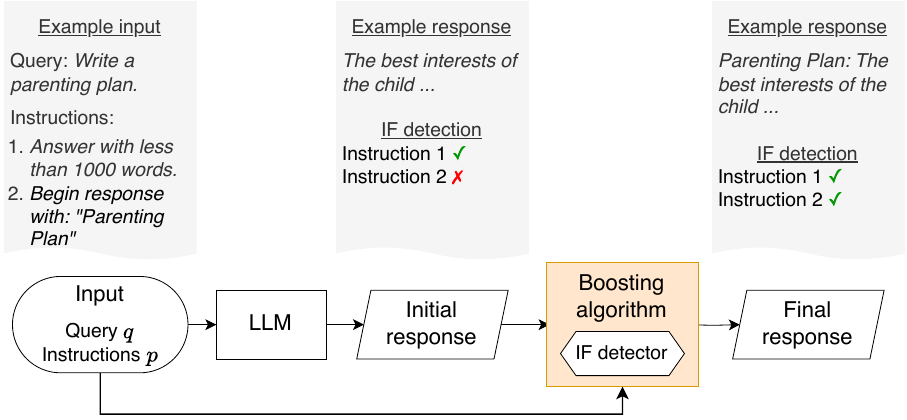}
    \caption{Overview of the instruction boosting approach.}
    \label{fig:overview}
\end{figure}

We propose \textit{Instruction Boosting} as a test-time post generation method to increase the reliability of LLM instruction following in applications. Instruction boosting is predicated on the observation that it is often easier for an LLM to revise a suboptimal response to meet a set of instructions than it is to generate a perfect response in the first place. As such, our method borrows from concepts of self-correction~\citep{madaan2023selfrefine, huang2024selfcorrect}, employing techniques to refine and reinforce initial model responses against the given instructions.
As shown in Fig.~\ref{fig:overview}, we apply boosting on an initial response from an LLM in order to increase the number of instructions that are followed. The boosting algorithm may internally use a detector to determine which instructions are being followed.
We evaluate several boosting strategies and show instruction following improvements across a range of open LLMs. Specifically, we show that boosting improves the IF rate by up to 7 percentage points in the case of 2 prompt instructions and by up to 4 percentage points for 10 instruction scenarios.  

To rigorously evaluate our approach, we contribute \textsc{ScaledIF}, a new instruction following benchmark that extends the popular IFEval dataset~\citep{zhou2023instruction} to include data samples with up to ten instructions. Our experimental results with \textsc{ScaledIF} also independently confirm the typical performance degradation with increasing instructions. Since each instruction constrains the model's response, many instructions can easily create an over-constrained problem. While some instruction sets may contain explicit conflicts—pairs of instructions that are impossible to follow simultaneously—we also show that even in the absence of such hard conflicts, a growing number of constraints can lead to tension between instructions, making it increasingly difficult to satisfy all of them at once.

We formalize this notion by defining a \textit{soft conflict} as a pair of instructions that are difficult, though not impossible, to follow simultaneously. We developed a quantitative conflict scoring test that determines the degree of soft conflict among a set of instructions. We show that conflict scores are negatively correlated with both the initial IF rate and the improved IF rate after boosting. 

We exploit this relationship by proposing the conflict scoring test as a valuable feedback tool for developers. Developers can compute conflict scores before and after adding additional instructions to a prompt to obtain crucial feedback about the impact the additional instructions have on model performance.  The conflict scoring test can serve as an early indicator of expected instruction following performance and can guide developers in adjusting or modifying instructions to lower the conflict score, thereby obtaining improved responses and better overall model control.

This paper makes the following contributions:
(i) the \textsc{ScaledIF} dataset with an instruction volume of up to 10 instructions per data sample,  
(ii) Instruction Boosting as a test-time method for developers to increase reliance on prompt-based instructions and control over LLM responses, and 
(iii) an instruction conflict scoring test  that estimates the complexity of satisfying a set of instructions simultaneously to provide feedback to developers.


\section{\textsc{ScaledIF} Dataset} \label{sec:dataset}

To study the effects that additional instructions have on the overall IF rate, we created \textsc{ScaledIF},  a derivative of the popular IFEval \citep{zhou2023instruction} instruction following dataset with up to ten instructions per data sample. The original IFEval dataset contains 541 samples, where each sample contains a query and between one and three verifiable instructions that must be followed. There are 26 distinct classes of instructions, each with its own Python verifier function to validate if a given text follows the instruction. 

An instruction is defined by a tuple: $(description,$ $class id$, $parameters)$, where $description$ is a text description (e.g, ``Answer with at least 50 words."),  $class id$ refers to an associated verifier function (e.g., {\em length\_constraints:number\_words}), and 
$parameters$ are actual verifier function parameters matching the instruction description (e.g., \{ relation: ``at least", num\_words: 50 \})

An instruction can be verified by invoking the associated verifier function with the given parameters. 

{\bf \textsc{ScaledIF}:} Our derivative dataset consists of 538 of the 541 IFEval samples, each containing a query (e.g.,  \textit{`Can you elaborate on ``I froze when I was jogging"?'}), and ten unique instructions. We rewrote the IFEval samples to isolate the query and remove existing instructions. 
We used Mistral Large~\citep{mistrallarge} for an initial extraction of the query and then refined the extracted queries manually. 

We added ten unique instructions to each sample. We leverage the fact that the 26 instruction classes are largely independent of the query. Any query can be paired with instructions from any unique instruction class. We made some changes to the instructions to ensure query-independence: we replaced three instruction classes (\textit{response\_language, english\_uppercase, english\_lowercase}) with three new query-independent ones (\textit{length\_constraints:sentence\_length, detectable\_format:yaml\_format, startend:start\_checker)}. This resulted in a set $\mathcal{P}$ of 26 query-independent instruction classes grouped into eight instruction categories: change case, combination, detectable content, detectable format, keywords, length constraints, punctuation, and startend. 

Each of the 538 samples was then assigned a set of $N=10$ instructions drawn from $\mathcal{P}$. Each instruction class $p\in \mathcal{P}$ is associated with a sampling function $S$ which samples values for its parameters, if any. For numerical parameters such as number of words or number of paragraphs, we sample parameter values from a simple 1-dimensional distributions. For parameters requiring keywords (e.g., \textit{`don't use the word ``hot"')}, keywords are sampled using an LLM (Granite 3.1-8B~\citep{granite2024}) in order to generate keywords that are relevant and meaningful in the context of a given query. This keyword sampling approach is in contrast to the original IFEval selection method which chose keywords uniformly at random from a large vocabulary.

\begin{algorithm}
\caption{Constrained Instruction Sampling}
\label{alg:sample_policies}
\begin{algorithmic}[1]

\REQUIRE Integer $N > 0$, set of Instructions $\mathcal{P}$
\ENSURE List $L$ of $N$ valid (Instruction, Parameters) samples

\STATE $L \leftarrow$ empty list

\WHILE{$\text{length}(L) < N$}
    \STATE Sample $p$ uniformly without replacement from $\mathcal{P}$
    \STATE Compute constraints $C_p$ on $p$ from instructions in $L$
    \STATE Sample required arguments $A_p$ using strategy $P(C_p)$
    \IF{a valid $A_p$ can be found}
        \STATE Append ($p$, $A_p$) to $L$
    \ENDIF
\ENDWHILE

\RETURN $L$

\end{algorithmic}
\end{algorithm}

\begin{figure}[t]
    \centering
    \includegraphics[width=1.0\linewidth]{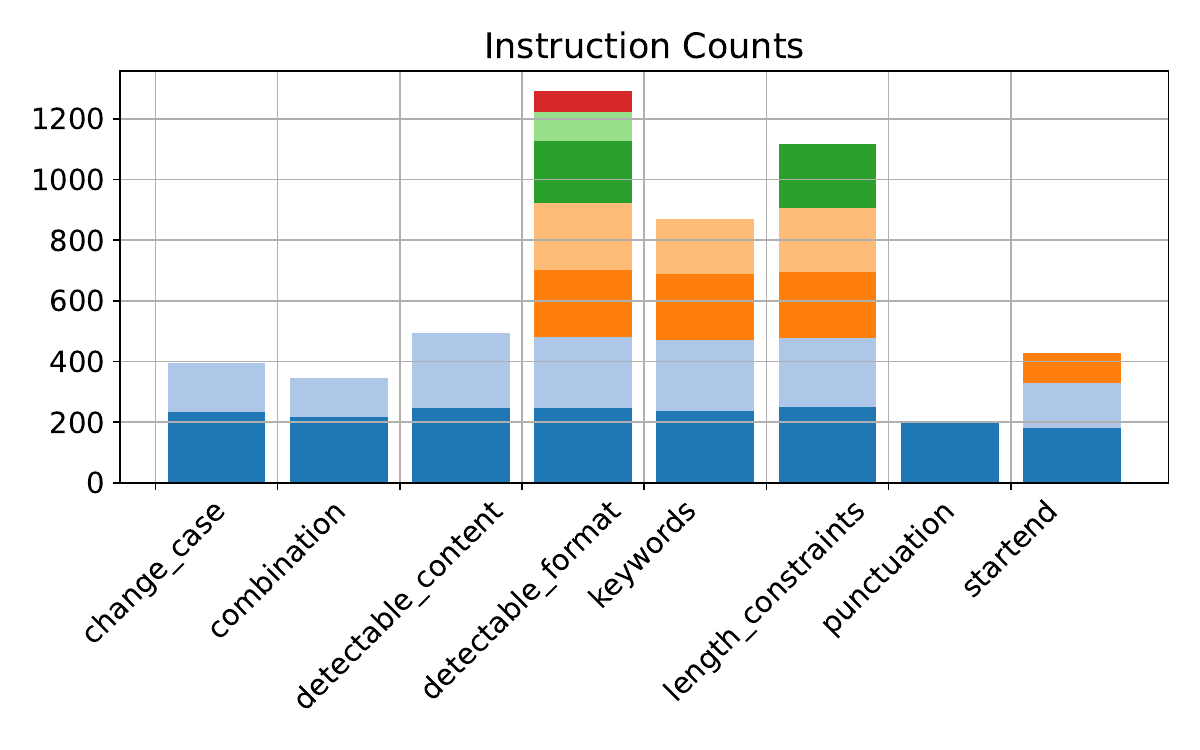}
    \caption{Number of instructions in \textsc{ScaledIF} across eight instruction categories. The contribution of individual instruction classes in each category is distinguished by color.}
    \label{fig:instructions}
\end{figure}

Fig.~\ref{fig:instructions} shows the number of instructions per instruction category in \textsc{ScaledIF}. A full list of the instruction classes is included in the Appendix. 

Sampling instruction parameters independently for each the $N$ instructions can easily lead to contradictions. For example, the {\em keywords:existence} instruction class requires that a set of keywords be present in the response, and the {\em keywords:forbidden\_words} instruction class requires that a set of keywords not appear. These parameters must be sampled to be disjoint to avoid contradictory instructions that are impossible to follow simultaneously.   

To solve this problem we enforced pairwise constraints during parameter sampling. When constructing an instruction set $L$ and a candidate instruction $I$ is selected to be added to $L$, first constraints $C_I$ are computed based on the instructions already chosen for $L$. For example, if an instruction stating that the response must contain at least $N$ paragraphs is already in $L$, then a constraint is added to ensure that no other instruction can require the number of words to be less than $N * 10$. These constraints are enforced while sampling parameters for the candidate instructions, and if they cannot be satisfied, then that instruction is rejected and a new candidate is sampled. If instruction parameters are successfully sampled which satisfy all existing constraints, then the candidate instruction is added to the list $L$. This process is described in Alg. \ref{alg:sample_policies}. 

After sampling ten instructions for each of the 538 samples, we shuffled the instruction order to avoid any sensitivity to the order in which instructions appear in the prompt. Finally, we constructed scaled down versions of the dataset with 2, 4, 6 and 8 instructions per sample by randomly removing instructions from the 10-instruction dataset version.


\section{Instruction Boosting} \label{sec:boosting}

The idea behind instruction boosting is to scale compute~\citep{snell2024scaling}
in order to improve on the baseline IF rate. As illustrated in Fig.~\ref{fig:overview}, instruction boosting operates as 
a post-generation step that, similar to  self-correction
~\citep{madaan2023selfrefine, huang2024selfcorrect}, employs techniques to refine and reinforce an initial model response against the given instructions.  
Instruction boosting can be enabled for a subset or all the instructions in given prompt. We devised several boosting strategies with different cost-performance trade-offs to give developers choice when balancing costs and benefits.

We evaluated instruction boosting on \textsc{ScaledIF} with several open models of various sizes including Llama-3.3-70B-Instruct\footnote{\cite{llama3.3-70b-instruct-hf}: Meta-Llama-3.3-70B-Instruct} (Llama-70B), Llama-3.1-8B-Instruct\footnote{\cite{llama3.1-8b-instruct-hf}: Meta-Llama-3.1-8B-Instruct} (Llama-8B), Qwen2.5-72B-Instruct\footnote{\cite{qwen2.5-72b-instruct-hf}: Qwen2.5-72B-Instruct} (Qwen-72B), Mixtral-8x7B-Instruct-v0.1\footnote{\cite{mixtral-8x7b-instruct-hf}: Mixtral-8x7B-Instruct-v0.1} (Mixtral-8x7B) and Mixtral-8x22B-Instruct-v0.1\footnote{\cite{mixtral-8x22b-instruct-hf}: Mixtral-8x22B-Instruct-v0.1} (Mixtral-8x22B).  

\begin{figure}[t]
    \centering
    \includegraphics[width=1.0\linewidth]{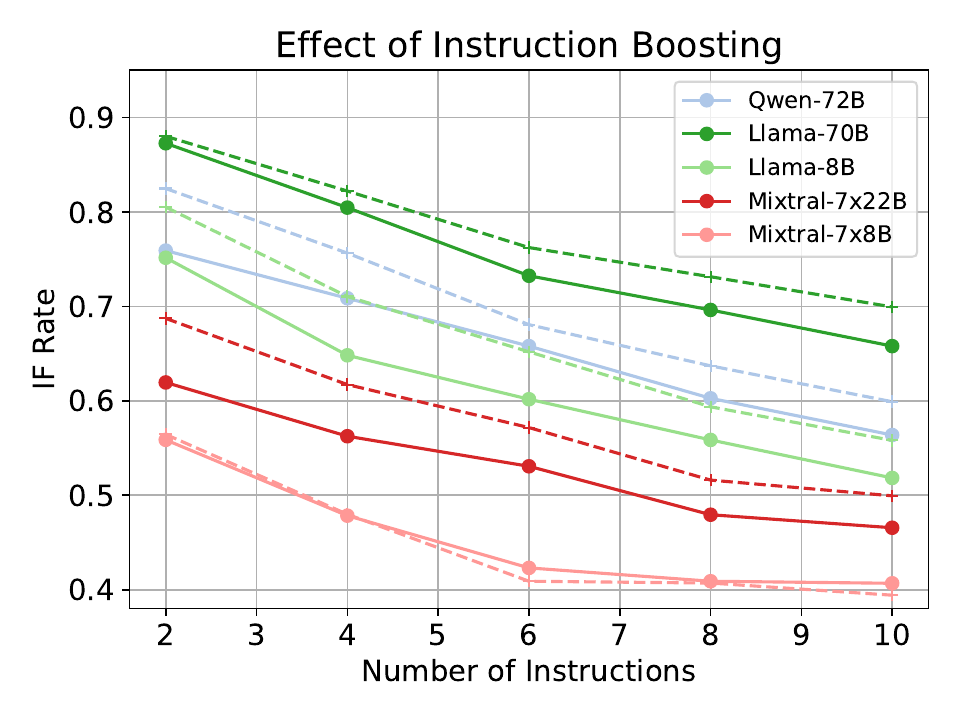}
    \caption{Initial IF rate (solid lines) and Best-of-N boosting performance (dashed lines).}
    \label{fig:modelcompare}
\end{figure}

As shown in Fig.~\ref{fig:modelcompare} (solid lines), the initial IF rate that these models achieve on \textsc{ScaledIF} ranges from 0.56 (Mixtral-7x8B) up to 0.88 (Llama-70B) with two instructions, and reduces to 0.39 (Mixtral-7x8B) and 0.66 (Llama-70B) with ten instructions. Fig.~\ref{fig:modelcompare} also shows the instruction following boost (dotted lines) achieved with the best performance boosting strategy, which will be explained in detail in the following section. With two instructions the largest IF rate boost is achieved by Mixtral-7x22B at 7 percentage points. With ten instructions, the largest boost is by Llama-70B at 4 percentage points. 

Note that across all models the IF rate drops as the number of instructions is increased, confirming previously observed instruction following degradation at scale. We will return to examine the factors that contribute to these degradation trends in Section~\ref{sec:conflict}.

\subsection{Boosting Strategies}

Instruction boosting is a test-time post-generation improvement strategy. A boosting strategy takes as input a query, a set of instructions and a generated response and produces as output a new response that has been revised to maximize adherence to the instructions. We devised the following boosting strategies (see Appendix for the corresponding prompts). 

{\bf Detect+Repair:}  Detect+Repair proceeds in two steps.  First, an LLM-as-a-judge detector (judge detector) is used to determine which instructions have not been followed in the input response. In the second repair step the response is rewritten to repair all detected instruction violations.   

{\bf Best-of-N}: Best-of-N samples $N$ rewritten responses that are to follow all instructions not already adhered to in the initial response using temperature sampling. Best-of-N does not require an initial detection step. Instead, the judge detector is used as a reward model to assign an instruction following reward to each rewritten response. The reward is the IF rate detected by the judge detector. In a final selection step, the response with the highest reward is chosen as the repaired response. In all experiments we used $N=5$ as we observed diminishing returns at higher sampling rates. 

{\bf Best-of-N Oracle}: To understand the potential IF rates models can achieve in their rewrites, we include a variant of Best-of-N that uses an oracle reward model to assign the actual instruction following rate to a given response. We used the deterministic IFEval instruction verifiers as the oracle. 

{\bf Map Reduce}: Map Reduce proceeds in three phases. First, the judge detector is used to detect violated instructions in the initial response. The Map phase creates separate rewrite tasks for each detected instruction violation. The final Reduce phase merges the independently generated rewritten responses into one final repaired response. 

\begin{figure}[t]
    \centering
    \includegraphics[width=1.0\linewidth]{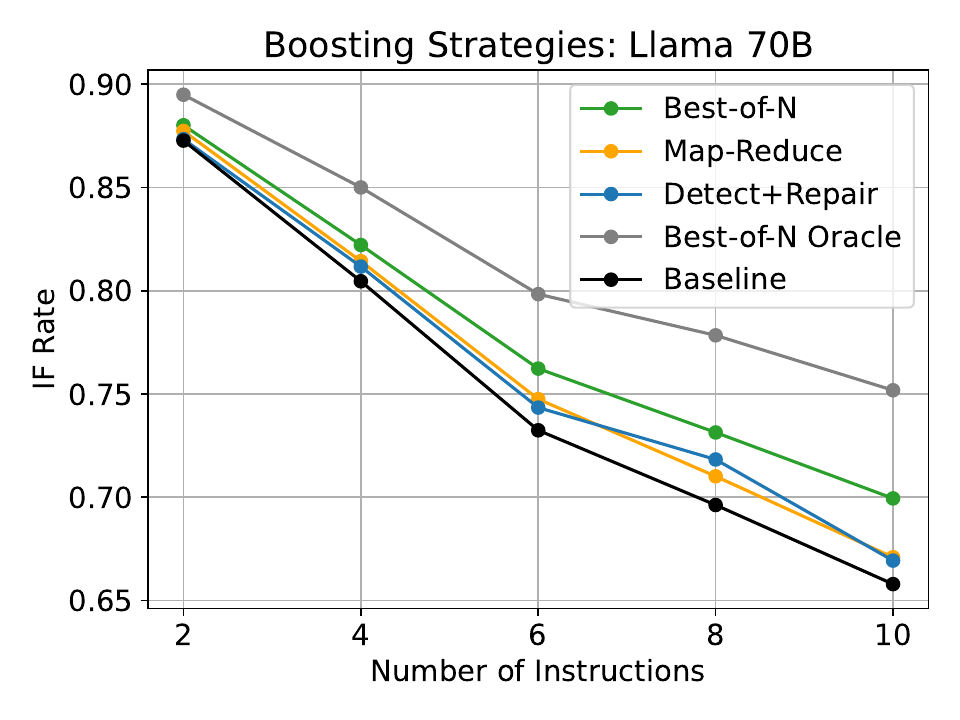}
    \caption{Instruction following rate achieved by Llama 70b for different boosting strategies.}
    \label{fig:strategiescompare}
\end{figure}

Fig.~\ref{fig:strategiescompare} shows the achieved IF rates for the four strategies for Llama-70B, the best performing model from Fig.~\ref{fig:modelcompare}. The IF rate achieved in the initial responses, prior to boosting is shown as the baseline. All boosting strategies lead to IF rate improvements over the baseline. Even at two instructions, boosting leads to small improvements and the benefits generally increase with the number of instructions. 

Among the non-oracle strategies, Best-of-N consistently provides the largest boost, up to 4 percentage points at ten instructions, increasing the IF rate to 0.70 from 0.66. Best-of-N Oracle shows the potential IF rate achievable through rewrite sampling.  Even at two instructions, the model is capable of generating rewritten responses with an IF rate of 0.89, a 2 percentage point increase. The boost grows as instructions are increased to ten, when the IF rate reaches 0.75, an 8.5 percentage point increase. Although not shown for space constraints, the relative boosting trends across the fours strategies are similar across all models from Fig.\ref{fig:modelcompare}.  

The gap between Best-of-N and Best-of-N Oracle is a result of inaccuracies in the judge detector that is used as the reward model. When using Llama-70B as the reward model, the detection accuracy is 73\%. Thus,  closing this gap may be achieved by replacing the judge detector with manually coded or LLM generated deterministic verifiers.

%

{\bf Task Adherence:} Since the instructions are mostly orthogonal to the query in each sample, it is possible to satisfy them while completely ignoring the primary task, namely answering the query. 
As a quality check, we used Llama-70B as a task-adherence judge on the initial model response and on the response after instruction boosting. This task-adherence judge was instructed to determine if the given response was related to the query (see Appendix for judge prompt). In the case that the initial response fails this check, the data point is not included in the results. If the initial response passed but after instruction boosting the response failed, this should be considered an additional failure mode. Table \ref{tab:task_adherence} shows that the largest number of task adherence failures in the initial response generation was incurred for 8 instructions with 22 out of 538 (4\%) failures. Boosting caused at most 7 (1.3\%) additional task adherence failures in the 10 instruction experiment.

\begin{table}[ht] \footnotesize
\centering
\begin{tabular}{|r||c|c|c|c|c|}
\hline
{\bf Num Instructions} & \textbf{2} & \textbf{4} & \textbf{6} & \textbf{8} & \textbf{10} \\
\hline
Initial Failures & 0 & 11  & 8 & 22 & 20 \\
Add'l after boosting & 0 & 0 & 6 & 3 & 7 \\
\hline
\end{tabular}
\caption{Task Adherence Failures (Llama-70B)}
\label{tab:task_adherence}
\end{table}

\begin{figure}[t]
    \centering
    \includegraphics[width=0.88\linewidth]{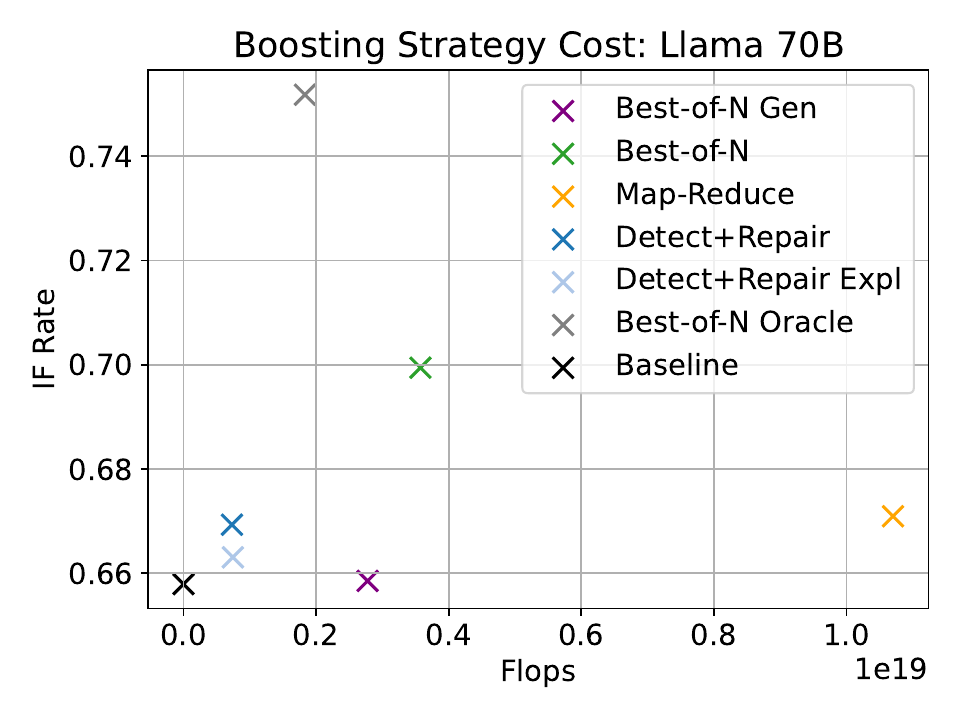}
    \caption{Costs (completion tokens in Flops) and IF rate of each strategy for 10 instructions achieved by Llama-70B.}
    \label{fig:costs}
\end{figure}

\subsection{Cost Analysis} 
In addition to the varying benefits among the boosting strategies we considered, they also incur different cost-benefit tradeoffs. To illustrate these tradeoffs Fig.~\ref{fig:costs} plots the achieved IF rate against cost measured as completion tokens in Flops. Compared to the lowest cost Detect+Repair strategy, Best-of-N trades additional compute for an additional IF rate boost. Map Reduce is less cost effective, requiring significantly more compute for a small IF rate increase. 

We also explored the following boosting variations. 

{\bf Detect+Repair Expl:} As a variant of Detect+Repair, we added explanations of instruction violations to the rewrite prompt. We expected the additional explanation hints to help the model in the rewrite task. Surprisingly, adding explanations didn't help and may have only provided a distraction to the model since they actually led to a small erosion of the IF rate improvements as shown in Fig.~\ref{fig:costs}.

{\bf Best-of-N Gen:} Instead of sampling $N$ response rewrites,  Best-of-N Gen samples $N$ initial response generations to the query. Both Best-of-N and Best-of-N Gen use the judge detector as a reward model.  As shown in the Fig.~\ref{fig:costs}, compared to rewrite sampling, sampling the initial responses incurs slightly lower cost but was not able to match the IF rate improvements of Best-of-N confirming our hypothesis that rewriting a draft response is generally easier than writing a response from scratch. 


\section{Instruction Scaling and Conflict Analysis}
\label{sec:conflict}

This section takes a closer look at the drivers for the observed instruction scaling trends that show decreasing IF rates with increasing numbers of instructions.

\subsection{Soft Conflicts}

In Section~\ref{sec:dataset}, we described how we applied pre-defined constraints to avoid contradictory instructions.
For example, an linstruction that requires the keyword `confidential' to appear in the response would contradict one that prohibits the same word.
We refer to a pair of contradictory instructions as a \textit{hard conflict}. Even after applying these pre-defined constraints and ruling out hard conflicts, there may still be instructions that are difficult for a model to follow simultaneously. For example, instructing a model to include at least 300 words in a response and also instructing it to not repeat any words may be difficult. We carry out a self-play approach to empirically identify such \textit{soft conflicts}.

\begin{figure}[tb]
    \centering
    \includegraphics[width=1.0\linewidth]{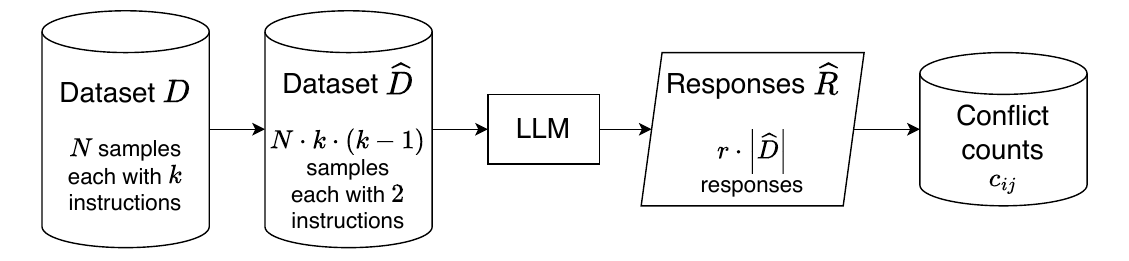}
    \caption{Steps to empirically compute an estimated conflict score between a pair of instructions.}
    \label{fig:conflict-steps}
\end{figure}

To quantify the degree of soft conflict in a sample, we compute a \textit{conflict score} between every pair of instructions assigned to that sample. The conflict score of a pair of instructions indicates the difficulty of following both instructions simultaneously. 
Conflict scores are computed by sampling multiple responses from a model for each instruction pair. Each response is checked to determine which instructions were followed: either both instructions were followed, or at least one was not followed. The latter case indicates possible soft conflicts. This flow is depicted in Figure~\ref{fig:conflict-steps}.

More precisely, we start from the original dataset $D$, outlined in Section~\ref{sec:dataset}, with $N$ samples, where each sample consists of a query and $k$ instructions. Next we construct a \emph{pairwise} dataset $\hat{D}$ as follows: for each sample in $D$ we create $k(k-1)$ additional samples with the same query but every subset of two instructions from the original $k$ instructions.
The model generates $r$ responses for each sample in $\hat{D}$ to produce the response set $\hat{R}$. The conflict count $c_{ij}$ between instructions $i$ and $j$ is  the number of responses in $\hat{R}$ where at least one of the two instructions $i$ or $j$ were not followed.

We can now use the pair-wise conflict counts to compute a conflict score for each sample $s \in D$. Let $p(s)$ be the set of instructions associated with sample $s$. The conflict score $c_s$ of sample $s$ is the normalized sum of the conflict counts for each pair of instructions in $s$:

\begin{equation}
    c_s = \frac
    {\sum_{(i,j) \in p(s) \times p(s), i \neq j} c_{ij}}
    {|p(s)|}.
\end{equation}

\begin{table}[tb] \footnotesize \centering
\begin{tabular}{|r||c|c|c|c|c|}
\hline
\textbf{Num Instructions} & \textbf{2} & \textbf{4} & \textbf{6} & \textbf{8} & \textbf{10} \\
\hline
Avg conflict score & 0.24 & 0.67 & 1.17 & 1.59 & 2.03 \\
Correl (after boosting) & -0.79 & -0.63 & -0.46 & -0.42 & -0.37 \\
\hline
\end{tabular}
\caption{Estimated conflict scores for Best-of-N boosting with Llama-70B.}
\label{tbl:conflict_scores}
\end{table}

{\bf Conflict score at scale:} We expect that samples with higher conflict scores include instructions that are more difficult for a model to follow and boost. We see this in Table~\ref{tbl:conflict_scores}, where the average conflict scores increase with the number of instructions in the dataset. This suggests that part of the reason that instruction following compliance rate decreases with the number of instructions, as observed in Section~\ref{sec:boosting}, is that there are inherent conflicts when more instructions are added. Recall that the conflict scores are computed pair-wise so the conflict scores themselves are not susceptible to instruction scaling effects.

{\bf Correlation of conflict score:} Table~\ref{tbl:conflict_scores} also lists the Pearson correlation between the IF rate and the conflict scores. The correlation decreases with increasing instructions; in the case with 10 instructions, the correlation is about -0.37.

We posit two reasons for decreasing correlation. First, our conflict scores are only based on pair-wise instruction conflicts. There may be more complex soft conflicts that only arise in the interplay of more than two instructions. For example, consider a set of three instructions that place limits on the number of words, the number of sentences, and the number of words per sentence in the response. Following any pair of these instructions is easier than following all three.
Another probable reason for the decreasing conflict correlation is that IF rates degrade with more instructions not only due to soft conflicts among the instructions, but also because there is still some effect of models struggling to follow increasing number of instructions.
Further investigating and quantifying the contributions of these possible causes is an avenue for future work.

\begin{figure}[tb]
    \centering
    \includegraphics[width=.88\linewidth]{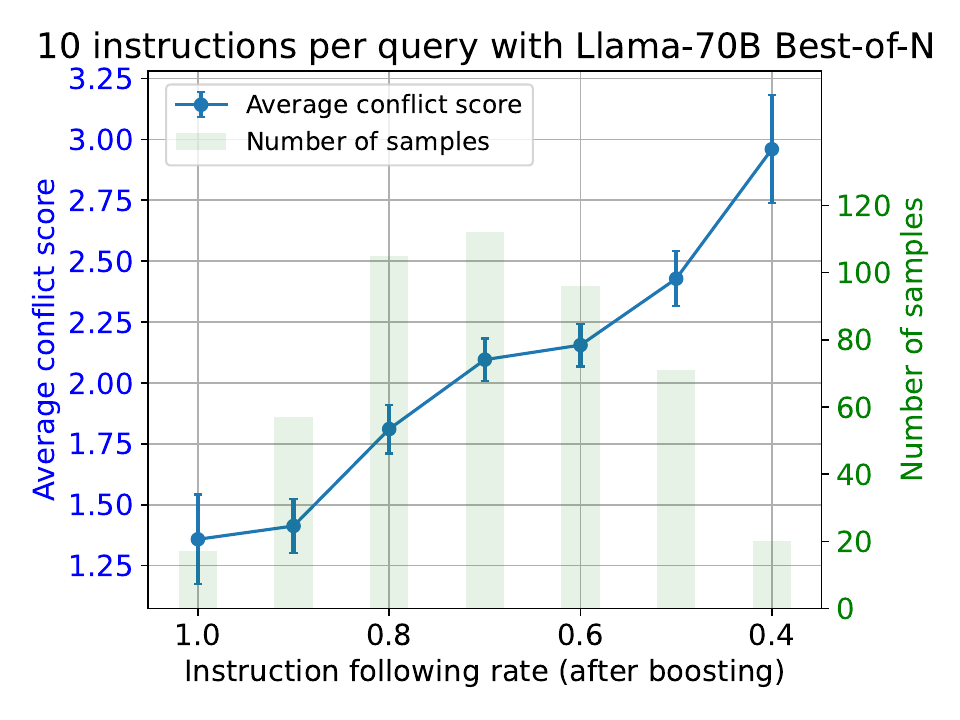}
    \caption{Average conflict scores for 10 instruction dataset, bucketed by the IF rate after boosting.
    }
    \label{fig:conflict_histogram}
\end{figure}

{\bf Segmentation by IF rate:} Let us dig deeper into distributions of the conflict scores. Figure~\ref{fig:conflict_histogram} plots the average conflict score for samples bucketed by the IF rate for the dataset with ten instructions. We see that ``harder'' samples, i.e., those with a lower IF rate, do indeed have a higher conflict score. This reinforces our assertion that soft conflicts among instructions contribute to worse IF rates when there are many instructions. Note that the error bars are the standard error of the mean, which is likely an underestimate of the true uncertainty due to correlations between instruction pairs appearing in multiple samples. 

{\bf Comparison without constraints:} Recall from Section~\ref{sec:dataset} that we apply pre-defined parameter sampling constraints to avoid including conflicting instructions in the dataset. We repeated the above analysis on a dataset where we did not apply these pre-defined constraints. As expected we observed a higher average conflict score: with 10 instructions the average conflict score was 18\% higher.



\subsection{Lost-in-the-Middle}

Prior work \citep{liu2023lost} looked into the ``lost-in-the-middle" effect and observed that performance is often highest when relevant information occurs at the beginning or end of the LLM input context, and significantly degrades when models must access relevant information in the middle.  

We investigated whether the lost-in-the-middle effect could play a role in the lowering IF rates we observed at rising numbers of instructions. Large numbers of instructions also have larger numbers of "middle" instructions.  To investigate, we broke down the IF rate by instruction position to compute positional IF rates as the n-th position IF rate. However, we found no consistent relationship between IF rates and instruction position across models. Middle instructions generally did not have lower IF rates than first or last instructions. Thus, the larger number of "middle" instructions in a list of 10 instructions does not appear to be a driver for the IF rate degradation we observed.

\section{Related Work}

Instruction following is critical to LLMs, enabling them to align with human intent. 
Related work on evaluating a model's instruction following ability includes IFEval~\citep{zhou2023instruction} which focuses on verifiable instructions that can be checked deterministically. IFEval includes data samples with one to three instructions per sample. \textsc{ScaledIF} extends IFEval by scaling instruction volume up to ten instructions per data sample. 
ComplexBench~\citep{wen2024complexbench} and FollowBench~\citep{jiang2024followbench} evaluate how LLMs handle complex instructions with different types of constraints and their compositions. 
InFoBench~\citep{qin2024infobench} introduces a new metric (DRFR) that breaks down complex instructions into simpler criteria for a more detailed analysis of a model's compliance with each part of the instruction. RefuteBench~\citep{yan2024refutebench} focuses on evaluating whether a model can modify its response to comply with human feedback in the form of refuting instructions in a conversational context. Verbalizer manipulation~\citep{li2024verbalizermanipulation} introduces an instruction following evaluation protocol for classification tasks. IFScale~\citep{jaroslawicz2025IFscale} is a benchmark of 500 keyword-inclusion instructions for a business report writing task and LIFBench`\citep{wu2025lifbenchevaluatinginstructionfollowing} focuses on evaluating instruction following in long context scenarios. 

Beyond evaluation, prior research has explored test-time interventions to improve reasoning and overall response quality. An approach similar to Detect+Repair boosting is self-correction~\citep{madaan2023selfrefine,huang2024selfcorrect} where a model is prompted to evaluate its own initial output and refine it. Self-correction primarily focuses on improving reasoning quality, whereas instruction boosting specifically targets a model's instruction following ability. 

Although not specifically aimed at instruction following,  Chain-of-Thought (CoT) prompting~\citep{wei2023chainofthought} can improve model performance on complex tasks by breaking down a difficult problem into smaller, more manageable steps.  Self-Consistency~\citep{wang2023selfconsistency} builds on CoT prompting by sampling multiple model responses and selecting the most frequent or ``consistent" response. Similar to Best-of-N boosting, self-consistency leverages a model's inherent capabilities to find a robust solution through compute scaling~\citep{snell2024scaling}. However, in contrast to self-consistency which samples response generation with the goal to improve CoT reasoning, Best-of-N boosting samples response rewrites and is targeted at improving IF rates. 

\cite{wu2024thinkingllmsgeneralinstruction} propose a training method for instruction following with thought generation and \cite{hou2025instructionfollowingpruninglargelanguage} propose a dynamic pruning approach to improve instruction following. In contrast, instruction boosting aims to improve instruction following as a post-processing method.

\section{Conclusion}

In this work, we've demonstrated that prompt-based instruction following in LLMs is a challenging problem, exacerbated by the instruction scaling effect. We introduced Instruction Boosting, a test-time strategy to enhance instruction adherence by refining and correcting initial model responses. Our empirical results on the \textsc{ScaledIF} dataset confirm that instruction boosting consistently improves instruction following, with gains of up to 7 and 4  percentage points for 2 and 10 instructions, respectively. We further contributed the concept of soft conflicts and a quantitative conflict score as a quantitative diagnostic tool for developers. By providing clear feedback to anticipate instruction following difficulty, our approach offers both a powerful method for improving model reliability and a valuable feedback mechanism for more effective prompt engineering and control over model behavior in applications. 

\bibliographystyle{aaai}
\bibliography{main}

\appendix

\section{\textsc{ScaledIF} Construction}

The following 26 instruction classes were used in the construction of the \textsc{ScaledIF} dataset. 

\begin{table}[ht] \footnotesize
\centering
\begin{tabular}{|r||l|}
\hline
 \multirow{5}{*}{length\_constraints}  & sentence\_length \\
 & number\_sentences \\
 & number\_words \\
 & number\_paragraphs \\
 & nth\_paragraph\_first\_word \\
 \hline
 \multirow{7}{*}{detectable\_format} & number\_bullet\_lists \\
 & constrained\_response \\
 & number\_highlighted\_sections \\
 &  title \\
 &  multiple\_sections \\
 &  json\_format \\
 &  yaml\_format \\
 \hline
 \multirow{2}{*}{detectable\_content}  &  number\_placeholders \\
  &  postscript \\
 \hline
 punctuation  &  no\_comma \\
 \hline
 \multirow{4}{*}{keywords}  &  forbidden\_words \\
  &  frequency \\
  &  letter\_frequency \\
  &  existence \\
 \hline
 \multirow{2}{*}{combination}  &  repeat\_prompt \\
  &  multiple\_responses \\
 \hline
 \multirow{2}{*}{change\_case}  &  capital\_word\_frequency \\
 &  lowercase\_sentences \\
 \hline
 \multirow{3}{*}{startend} &  quotation \\
 &  start\_checker \\
 &  end\_checker \\
\hline
\end{tabular}
\caption{\textsc{ScaledIF} Instructions}
\label{tab:scaledif-instructions}
\end{table}
All but three (length\_constraints:sentence\_length, detectable\_format:yaml\_format, startend:start\_checker) of these instructions were part of the original IFEval dataset. In addition, three of the original IFEval instructions (response\_language, english\_uppercase, english\_lowercase) with three new query-independent ones (length\_constraints:sentence\_length, detectable\_format:yaml\_format, startend:start\_checker) were removed because they were too difficult to combine with other instructions for scaling. 

\subsection{Constraints}

In order to ensure that the sets of instructions constructed for \textsc{ScaledIF} are mutually satisfiable, constraints had to be imposed on the sampled parameters. For example suppose that length\_constraints:number\_paragraphs is sampled with parameters num\_paragraphs = 3, and then length\_constraints:number\_words is sampled. Before sampling the parameters for length\_constraints:number\_words, all constraints related to the previously sampled instructions will be computed. The length\_constraints:number\_paragraphs instruction will produce the constraint num\_words $>= 10*$ num\_paragraphs $= 30$. This constraint is chosen to ensure that the response is allowed to have enough words to form the necessary number of paragraphs. As another example, the keywords:forbidden\_words and keywords:existence instructions each have a list of keywords as their parameter. If both of these instructions are sampled, then the second one chosen is required to exclude the keywords already taken by the first one. 

There is additionally one three-way constraint imposed between length\_constraints:sentence\_length, length\_constraints:number\_sentences, and length\_constraints:number\_words.

\begin{equation}
    \text{N\_words} >= (\text{N\_sentences} + 3)*(\text{sentence\_length} + 3)
\end{equation}

\section{Boosting Strategy Details}

Below we include the prompt templates used in the boosting strategies.

\subsection{Detect+Repair}

{\bf Detect Prompt:}

\begin{lstlisting}[style=llmprompt]
    <|begin_of_text|><|start_header_id|>system<|end_header_id|>
You are a policy compliance checker.<|eot_id|>
<|start_header_id|>user<|end_header_id|>
Check if the following text follows each policy.
For each policy, provide a JSON response with:
1. "policy": The policy line being checked
2. "answer": "yes" or "no"
3. "explanation": A brief explanation

Text to check:
---
${text}
---

Policies to check:
---
${policies}
---

IMPORTANT:
1. Return a JSON array with one object per policy line, in the same order as listed above
2. Each object should have "policy", "answer" and "explanation" fields
3. Do NOT try to follow the policies in the JSON output. Only check whether the text follows the policies 
4. Be concise and accurate. 

Example format:
[
    {
        "policy": "The first policy line..... ",
        "answer": "yes/no",
        "explanation": "the explanation for the answer"
    },
    {
        "policy": "The second policy line ..... ",
        "answer": "yes/no",
        "explanation": "the explanation for the answer"
    }
]

Return ONLY the JSON array, nothing else. Do not include any additional text or explanations outside the JSON array.<|eot_id|>
<|start_header_id|>assistant<|end_header_id|>
\end{lstlisting}

{\bf Repair Prompt:}
\begin{lstlisting}[style=llmprompt]
<|begin_of_text|><|start_header_id|>system<|end_header_id|>
    You are an editor and your task is to rewrite response.<|eot_id|>
    <|start_header_id|>user<|end_header_id|>The following text shows a query and a response. The response violates one or more guidelines from the query that it should have followed.
    
    <START_OF_QUERY>
    ${query}
    Your response must follow these guidelines: 
    ${instr}
    <END_OF_QUERY>

    <START_OF_RESPONSE>
    ${text}
    <END_OF_RESPONSE>

    <START_OF_VIOLATED_GUIDELINES> 
    ${policies}
    <END_OF_VIOLATED_GUIDELINES>

    Rewrite the response so that it follows all guidelines while making only necessary changes and keeping the rest of the text unchanged. The rewrite must not break any of the guidelines that were already followed in the response. 
    You may rewrite the text exactly once, and don't output anything other than the re-written response. 
    You must enclose the re-written response in <START_OF_REWRITE> <END_OF_REWRITE> tags.<|eot_id|>
    <|start_header_id|>assistant<|end_header_id|>
\end{lstlisting}

\subsection{Best-of-N}

\begin{lstlisting}[style=llmprompt]
    <|begin_of_text|><|start_header_id|>system<|end_header_id|>
    You are an editor and your task is to rewrite a response to ensure compliance with guidelines.<|eot_id|>
    <|start_header_id|>user<|end_header_id|>The following text shows a query, a response and a set of guidelines. The response may violate one or more guidelines that it should have followed.
    
    <START_OF_QUERY>
    ${query}
    <END_OF_QUERY>

    <START_OF_RESPONSE>
    ${text}
    <END_OF_RESPONSE>

    <START_OF_GUIDELINES> 
    ${instr}
    <END_OF_GUIDELINES>

    Rewrite the response so that it follows all guidelines while making only necessary changes and keeping the rest of the text unchanged. The rewrite must not break any of the guidelines that were already followed in the response. 
    You may rewrite the text exactly once, and don't output anything other than the re-written response. 
    You must enclose the re-written response in <START_OF_REWRITE> <END_OF_REWRITE> tags.<|eot_id|>
    <|start_header_id|>assistant<|end_header_id|>
\end{lstlisting}

\subsection{MapReduce}

\begin{lstlisting}[style=llmprompt]
    <|begin_of_text|><|start_header_id|>system<|end_header_id|>
You are an editor and your task is to generate a response according to a user query and to ensure compliance with a set of guidelines.<|eot_id|>

<|start_header_id|>user<|end_header_id|>The following text shows the user query, a set of per-guideline responses, and the set of guidelines. Each per-guideline response is an example of a response that complies with the associated guideline.

<START_OF_QUERY>
${query}
<END_OF_QUERY>
${per_guideline_responses}
<START_OF_GUIDELINES> 
${instr}
<END_OF_GUIDELINES>

Generate a response so that it follows all the guidelines. Use the per-guideline responses to help generate the final response that complies with all the guidelines.
Don't output anything other than the re-written response. 
You must enclose the re-written response in <START_OF_REWRITE> <END_OF_REWRITE> tags.<|eot_id|>
<|start_header_id|>assistant<|end_header_id|>
\end{lstlisting}

\subsection{Best-of-N Gen}

\begin{lstlisting}[style=llmprompt]
    <|begin_of_text|><|start_header_id|>system<|end_header_id|>
    You are a helpful assistant and your task is to respond to queries. Your response must follow a set of guidelines.<|eot_id|>
    <|start_header_id|>user<|end_header_id|>Here is a query with a set of guidelines that must be followed.
    
    <START_OF_QUERY>
    ${query}
    Your response must follow these guidelines: 
    ${instr}
    <END_OF_QUERY>

    Write a response that follows all guidelines and don't output anything other than the response. 
    You must enclose the response in <START_OF_RESPONSE> <END_OF_RESPONSE> tags.<|eot_id|>
    <|start_header_id|>assistant<|end_header_id|>
\end{lstlisting}

\section{Task Adherence Judge}

\begin{lstlisting}[style=llmprompt]
    <|begin_of_text|><|start_header_id|>system<|end_header_id|>
    You are a helpful assistant whose task is to respond to user queries.  
    Make sure that your response follows these instructions: 

    {instruction_list}<|eot_id|>
    <|start_header_id|>user<|end_header_id|>
    Question:
    {prompt_request}<|eot_id|>
    <|start_header_id|>assistant<|end_header_id|>
\end{lstlisting}

\end{document}